\documentclass[
twocolumn,
]{ceurart}

\usepackage{subfig}
\usepackage{graphicx}
\usepackage{multirow}
\usepackage{makecell}
\usepackage{enumitem}

\usepackage{listings}
\begin{document}

\copyrightyear{2024}
\copyrightclause{Copyright for this paper by its authors.
  Use permitted under Creative Commons License Attribution 4.0
  International (CC BY 4.0).}

\conference{IR-RAG@SIGIR' 24: The 47th International ACM SIGIR Conference on Research and Development in Information Retrieval, July 14--18, 2024, Washington D.C., USA}

\title{The Impact of Quantization on Retrieval-Augmented Generation: An Analysis of Small LLMs}



\author[1,2]{Mert Yazan}[%
orcid=0009-0004-3866-597X,
email=m.yazan@hva.nl,
]
\cormark[1]

\author[2]{Suzan Verberne}[%
orcid=0000-0002-9609-9505,
email=s.verberne@liacs.leidenuniv.nl,
]

\author[1]{Frederik Situmeang}[%
orcid=0000-0002-2156-2083,
email=f.b.i.situmeang@uva.nl,
]

\address[1]{Amsterdam University of Applied Sciences, Fraijlemaborg 133, 1102 CV Amsterdam, Netherlands}
\address[2]{Leiden University, Einsteinweg 55, 2333 CC Leiden, Netherlands}

\cortext[1]{Corresponding author.}

\begin{abstract}
    Post-training quantization reduces the computational demand of Large Language Models (LLMs) but can weaken some of their capabilities. Since LLM abilities emerge with scale, smaller LLMs are more sensitive to quantization. In this paper, we explore how quantization affects smaller LLMs' ability to perform retrieval-augmented generation (RAG), specifically in longer contexts. We chose personalization for evaluation because it is a challenging domain to perform using RAG as it requires long-context reasoning over multiple documents. We compare the original FP16 and the quantized INT4 performance of multiple 7B and 8B LLMs on two tasks while progressively increasing the number of retrieved documents to test how quantized models fare against longer contexts. To better understand the effect of retrieval, we evaluate three retrieval models in our experiments. Our findings reveal that if a 7B LLM performs the task well, quantization does not impair its performance and long-context reasoning capabilities. We conclude that it is possible to utilize RAG with quantized smaller LLMs. 
\end{abstract}


\maketitle

\section{Introduction}

    Large Language Model (LLM) outputs can be enhanced by fetching relevant documents via a retriever and adding them as context for the prompt. The LLM can generate an output grounded with relevant information with the added context. This process is called Retrieval Augmented Generation (RAG). RAG has many benefits such as improving effectiveness in downstream tasks \cite{Huang2023, Ma2023, Shi2023, Xu2023}, reducing hallucinations \cite{Proser2023}, increasing factuality \cite{Nakano2022}, by-passing knowledge cut-offs, and presenting proprietary data that is not available to the LLMs. 
    
    The performance of RAG depends on the number, quality, and relevance of the retrieved documents \cite{gao2024}. To perform RAG, many tasks demand a lot of passages extracted from multiple, unstructured documents: For question-answering tasks, the answer might be scattered around many documents because of ambiguity or the time-series nature of the question (eg. price change of a stock). For more open-ended tasks like personalization, many documents from different sources might be needed to capture the characteristics of the individual. Therefore to handle RAG in these tasks, an LLM needs to look at multiple sources, identify the relevant parts, and compose the most plausible answer \cite{gao2024}. 
    
    LLMs do not pay the same attention to their whole context windows, meaning the placement of documents in the prompt directly affects the final output \cite{Nelson2023}. On top of that, some of the retrieved documents may be unrelated to the task, or they may contain contradictory information compared to the parametric knowledge of the LLM \cite{xu2024knowledge}. An LLM has to overcome these challenges to leverage RAG to its advantage. \citet{Xu2023} have shown that an open-source 70B LLM \cite{Touvron2023} equipped with RAG can beat proprietary models, meaning it is not necessary to use an LLM in the caliber of GPT-4 \cite{OpenAI2023} to implement RAG. Still, for many use cases, it might not be feasible to deploy a 70B LLM as it is computationally demanding. To decrease the computational demand of LLMs, post-training quantization can be used. Quantization drastically reduces the required amount of RAM to load a model and can increase the inference speed by more than 3 times \cite{Dettmers2023, Frantar2023}. Despite the benefits, quantization affects LLMs differently depending on their size \cite{li2024}. For capabilities that are important to RAG, such as long-context reasoning, smaller LLMs (<13B) are found to be more sensitive to quantization \cite{li2024}. 

    \begin{figure*}[h]
    \centering
      \includegraphics[trim={0 16.5cm 0 0},clip,width=1\textwidth]{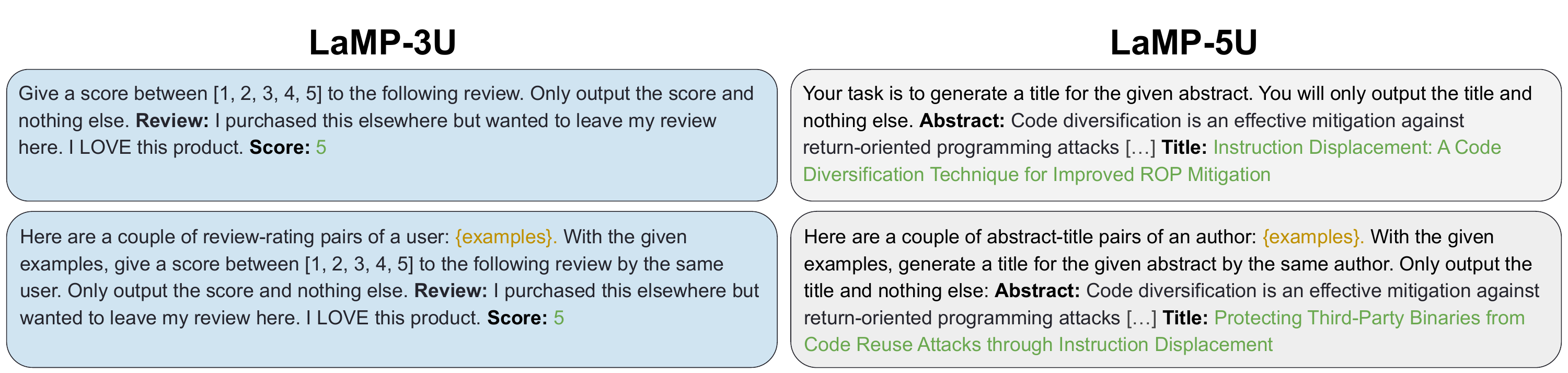}
      \caption{
        Prompts used for both datasets. The ones on the top represent $k=0$ (zero-shot, no retrieved documents) and the ones on the bottom are for $k>0$ settings (RAG). The green text is the model output. Line endings are not shown for space reasons.
       }
      \label{fig:prompt}
    \end{figure*}

    In this paper, we investigate the effectiveness of quantization on RAG-enhanced 7B and 8B LLMs. We evaluate the full (FP16) and quantized (INT4) versions of multiple LLMs on two personalization tasks taken from the LaMP \cite{Salemi2023} benchmark. To better study how quantized LLMs perform in longer contexts, we compared the performance gap between FP16 and INT4 models with an increasing number of retrieved documents. We chose personalization because it is a challenging task to perform with RAG as it demands long-context reasoning over many documents. Contrary to question-answering where the LLM has to find the correct answer from a couple of documents, personalization requires the LLM to carefully study a person's style from all the provided documents. Our findings show that the effect of quantization depends on the model and the task: we find almost no drop in performance for OpenChat while LLaMA2 seems to be more sensitive. Our experiments show that quantized smaller LLMs can be good candidates for RAG pipelines, especially if efficiency is essential.

\section{Approach}

    \subsection{LLMs}
    
    Starting with LLaMA2-7B (Chat-hf) \cite{Touvron2023} to have a baseline, we experiment with the following LLMs: LLaMA3-8B \cite{LLaMA3}, Zephyr (Beta) \cite{Tunstall2023}, OpenChat (3.5) \cite{Wang2023openchat}, and Starling (LM-alpha) \cite{Starling2023}. These models were chosen because they were the highest-ranked 7B and 8B LLMs in the Chatbot Arena Leaderboard \cite{Zheng2023} according to the Elo ratings at the time of writing. Since all models except LLaMA are finetuned variants of Mistral-7B \cite{Jiang2023}, we add Mistral-7B (Instruct-0.1)\footnote{Although there is an updated v0.2 version of Mistral-7B, we used v0.1 to match the other LLMs that are finetuned on it} to our experiments too. We use Activation-aware Weight Quantization (AWQ) as it outperforms other methods \cite{Lin2023}. 
    
    \subsection{Tasks and Datasets}
        
    We use the LaMP benchmark that offers 7 personalization datasets with either a classification or a generation task \cite{Salemi2023}. To represent both types of tasks, we chose one dataset from each: LaMP-3 (``Personalized Product Rating'') and LaMP-5 (``Personalized Scholarly Title Generation''). LaMP-3 is composed of product reviews and their corresponding scores. For each user, one of the review--score pairs is chosen as the target and other pairs become the user profile. The LLM's task, in this case, is to predict the score given a review using the other review--score pairs of the same user. LaMP-5 aims to generate a title for an academic paper based on the abstract. In this case, the user profile consists of abstract--title pairs that demonstrate the writing style of the user (scholar). The task of the LLM is to generate a title for the given abstract by incorporating the writing style of the scholar. Those datasets were chosen because compared to the other ones, on average, they had more samples in their user profiles, and the samples were longer. Therefore, they represented a better opportunity to evaluate RAG effectiveness as the retrieval part would be trickier.
    
    We work with the user-based splits (LaMP-3U, LaMP-5U) where the user appears only in one of the data splits \cite{Salemi2023}. The labels for the test sets are not publicly available (results can be obtained by submitting the predictions to the leaderboard) and since we did not fine-tune our models, we chose to use the validation sets for evaluation. For both datasets, we noticed that some samples do not fit in the context windows. After analyzing the overall length of the samples, we concluded that those cases only represent a tiny minority and removed data points that are not in the 0.995th percentile. For LaMP-5U, we also removed abstracts that consisted only of the text ``no abstract available''. There are 2500 samples in the validation sets, and we have 2487 samples left after the preprocessing steps for both datasets.

    \subsubsection{Evaluation} We used mean absolute error (MAE) for LaMP-3 and Rouge-L \cite{Lin2004} for LaMP-5, following the LaMP paper \cite{Salemi2023}. Their experiments also include root mean square error (RMSE) and Rouge-1 scores, but we found that the correlation between MAE and RMSE is 0.94, and between Rouge-1 and Rouge-L is 0.99. Therefore, we do not include those metrics in our results. The prompts we use are shown in Figure~\ref{fig:prompt}. Even though the LLMs are instructed to output only the score or the title, we notice that some are prone to give lengthy answers such as ``Sure, here is the title for the given abstract, Title: (generated title)''. We apply a post-processing step on the LLM outputs to extract only the score or the title before evaluation.  

    \subsection{Retrieval}
    
    We conduct the experiments with the following number of retrieved documents: $k \in \{0, 1, 3, 5, max\_4K, max\_8K\}$. $0$ refers to zero-shot without any retrieval and $max$ is the maximum number of documents that can be put into the prompt, given the context window of the LLM. LLaMA2 has a context window of 4096 tokens while other models have 8192 tokens. To make it fair, we include two options for the $max$ setting: 4K and 8K. For $max\_4K$, we assume that all models have a 4096 token context window, we use the original 8192 token context windows for $max\_8K$. Consequently, LLaMA2-7B is not included in the $max\_8K$ experiments. To put it into perspective, the number of retrieved documents varies between 15-18 for $max\_4K$, and between 25-28 for $max\_8K$ in LaMP-5U, depending on the average length of documents in the user profile. As retrievers, we evaluate BM25 \cite{Robertson1994} (BM25 Okapi) \footnote{\url{https://pypi.org/project/rank-bm25/}}, Contriever \cite{Izacard2022-2} (finetuned on MS-Marco), and DPR \cite{Karpukhin2020} (finetuned on Natural Questions)\footnote{\url{https://huggingface.co/facebook/dpr-question_encoder-single-nq-base}}. Since we focus on efficiency by reducing the computational load, the retrievers are not finetuned on the datasets.

    \begin{table*} 
        \scriptsize
        \caption{The absolute percentage change between FP16 and INT4 scores, using Contriever. More than a 5\% drop in performance is highlighted in red. For MAE, the lower is better while the inverse is true for Rouge-L.}
        \label{table:results-table}
        \begin{tabular}{c|c|c|ccccccccccccc}
        \multicolumn{3}{c}{} & \multicolumn{2}{c}{\textbf{LLaMA2}} & \multicolumn{2}{c}{\textbf{OpenChat}} & \multicolumn{2}{c}{\textbf{Starling}} & \multicolumn{2}{c}{\textbf{Zephyr}} & \multicolumn{2}{c}{\textbf{Mistral}} & \multicolumn{2}{c}{\textbf{LLaMA3}} \\
        \cmidrule(lr){4-5} \cmidrule(lr){6-7} \cmidrule(lr){8-9} \cmidrule(lr){10-11} \cmidrule(lr){12-13} \cmidrule(lr){14-15}
        \textbf{Dataset} & \textbf{Metric} & \textbf{k} & \textbf{FP16} & \textbf{INT4} & \textbf{FP16} & \textbf{INT4} & \textbf{FP16} & \textbf{INT4} & \textbf{FP16} & \textbf{INT4} & \textbf{FP16} & \textbf{INT4} & \textbf{FP16} & \textbf{INT4}\\
        \hline
        \multirow{6}{*}{LaMP-3U} & \multirow{6}{*}{MAE $\downarrow$} & 0 & 0.684 & +2.9\% & 0.440 & \textcolor{red}{-7.8\%} & 1.603 & +45\% & 0.435 & \textcolor{red}{-14.7\%} & 0.569 & -2.5\%  & 0.481 & \textcolor{red}{-5.9\%}\\
         & & 1 & 0.453 & -1.1\% & 0.312 & +5.5\% & 0.800 & +7.1\% & 0.300 & +1.9\% & 0.461 & \textcolor{red}{-9.3\%} & 0.364 & \textcolor{red}{-10.8\%} \\
         & & 3 & 0.637 & \textcolor{red}{-7.6\%} & 0.256 & +2.8\% & 0.718 & \textcolor{red}{-30.0\%} & 0.273 & +2.6\% & 0.404 & \textcolor{red}{-8.0\%} & 0.320 & \textcolor{red}{-9.2\%} \\
         & & 5 & 0.724 & \textcolor{red}{-23.3\%} & 0.238 & +1.8\% & 0.797 & \textcolor{red}{-32.0\%} & 0.266 & +0.8\% & 0.380 & \textcolor{red}{-8.1\%} & 0.305 & \textcolor{red}{-13.0\%}\\
         & & max\_4K & 0.508 & \textcolor{red}{-80.2\%} & \textbf{0.224} & +1.8\% & 0.985 & \textcolor{red}{-57.1\%} & 0.237 & -4.4\% & 0.346 & \textcolor{red}{-14.7\%} & 0.285 & \textcolor{red}{-23.1\%}\\
         & & max\_8K & - & - & 0.257 & -3.9\% & 1.352 & -1.1\% & 0.392 & \textcolor{red}{-6.3\%} & 0.368 & \textcolor{red}{-16.1\%} & 0.288 & \textcolor{red}{-23.4\%}\\
        \hline
        \multirow{6}{*}{LaMP-5U} & \multirow{6}{*}{Rouge-L $\uparrow$} & 0 & 0.338 & -0.6\% & 0.361 & -0.5\% & 0.359 & -2.3\% & 0.335 & -0.4\% & 0.361 & +1.2\% & 0.384 & -2.5\% \\
         & & 1 & 0.380 & \textcolor{red}{-9.7\%} & 0.404 & -1.0\% & 0.397 & -1.0\% & 0.360 & +0.9\% & 0.400 & -0.9\% & 0.402 & \textcolor{red}{-5.6\%}\\
         & & 3 & 0.385 & \textcolor{red}{-11.6\%} & 0.415 & 0.0\% & 0.412 & -0.2\% & 0.360 & -0.8\% & 0.410 & -0.5\% & 0.404	& \textcolor{red}{-5.2\%}\\
         & & 5 & 0.374 & \textcolor{red}{-10.3\%} & \textbf{0.422} & -0.7\% & 0.419 & -1.4\% & 0.365 & -2.1\% & 0.415 & -0.7\% & 0.397 & -4.5\%\\
         & & max\_4K & 0.337 & \textcolor{red}{-16.8\%} & 0.419 & -1.1\% & 0.402 & -0.5\% & 0.357 & \textcolor{red}{-7.7\%} & 0.410 & -1.1\% & 0.376 & -2.6\%\\
         & & max\_8K & - & - & 0.395 & -1.0\% & 0.379 & -1.9\% & 0.326 & \textcolor{red}{-18.7\%} & 0.387 & -0.8\% & 0.384 & \textcolor{red}{-7.2\%} \\
        \end{tabular}
        
    \end{table*}

\section{Results}

    \begin{figure*}[t]
    \subfloat[MAE results for LaMP-3U, the lower the better]{
      \includegraphics[width=0.98\textwidth]{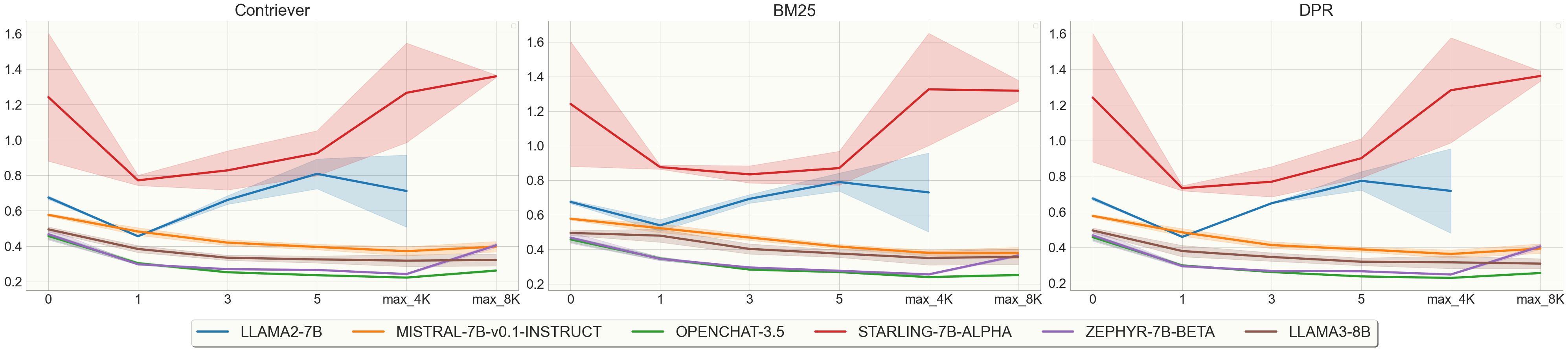}
      \label{fig:res_comp_3}
    }
    \\
    \subfloat[Rouge-L results for LaMP-5U, the higher the better]{
      \includegraphics[width=0.98\textwidth]{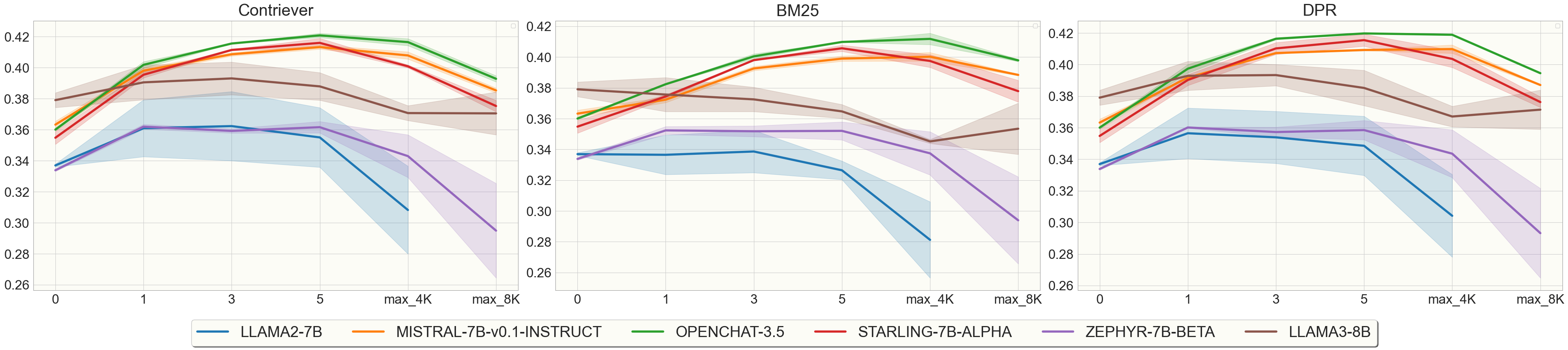}
      \label{fig:res_comp_5}
    }
    \caption{Results for both datasets. The upper and lower borders of each colored area represent the quantized and not-quantized performances of the models, and the corresponding lines are the mean of both.}
    \label{fig:res_comp}
    \end{figure*}
    
    \subsection{LLMs} The results are shown in Table~\ref{table:results-table}. We see the dominance of OpenChat in both datasets. Zephyr performs very close to OpenChat in LaMP-3U but falls far behind in LaMP-5U. The same can be said for Starling but reversed. Mistral-7B performs stable in both datasets albeit being slightly behind OpenChat. Overall, LLaMA2 performs the worst as it is below average in both datasets. Despite being the dominant small LLM currently, LLaMA3 is not the best for both tasks, despite performing reasonably well in LaMP-3U. Interestingly, LLaMA3 has the best zero-shot score in LaMP-5U but it struggles to improve itself with retrieval.
    
    \subsection{Quantization} How much an LLM gets affected by quantization seems to be related to how well it performs the task. OpenChat suffers almost no performance degradation from quantization. On the contrary, LLaMA2 seems very sensitive, especially when the number of retrieved documents is increased. Starling suffers no significant consequence from quantization in LaMP-5U where it performs well, but it does suffer in LaMP-3U. There also seems to be a disparity between the tasks as quantized LLMs perform much worse in LaMP-3U than in LaMP-5U. 
    
    \subsection{Number of retrieved documents} Figure~\ref{fig:res_comp} shows that LLM performance is saturated with a couple of documents, and the improvement obtained from more is marginal. In LaMP-5U, adding more than 5 documents starts to hurt the performance: Figure~\ref{fig:res_comp_5} shows an inverse-U-shaped distribution for all LLMs except LLaMA3. For some models, performance even drops below the zero-shot setting when all the available context window is filled with retrieved documents. The LaMP-3U results (Figure~\ref{fig:res_comp_3}) continue to improve after adding more than 5 documents as can be observed from $max\_4K$ scores, but MAE also starts to get worse for $max\_8K$.

    We analyze whether a longer context window hurts the quantized variants more and find that there seems to be a peculiar relationship. INT4 LLaMA2 suffers from longer contexts, while INT4 OpenChat performs well and acts almost the same as its FP16 counterpart. INT4 Mistral and LLaMA3 act very similar to their FP16 counterparts in LaMP-5U but in LaMP-3U, they get progressively worse with more documents. Overall, quantization can increase the risk of worsened long-context capabilities but there is not a direct relationship as it is highly task and context-dependent.

    \subsection{Retrievers}  Figure~\ref{fig:res_comp} shows that the three retrievers gave almost identical results, albeit BM25 being marginally behind the others. Also in LaMP-5U, the gap between the FP16 and INT4 LLaMA2 varies slightly.  Other than that, the retriever model does not have a noticeable impact on the personalization tasks we experimented with. The patterns we found regarding LLMs, quantization, and the number of retrieved documents are the same for all the retrievers. 
    
    \begin{table}[]
    \caption{Our results compared with LaMP. Results indicated with * are not significantly lower than the reported best result (FlanT5-XXL). A quantized 7B LLM can perform on par with a larger model while being much more efficient.
    }
    \label{table:lamp_comp}
    \begin{tabular}{l|c|c|c} 
     & \textbf{\makecell{LaMP-3U \\ (MAE) $\downarrow$ }} & \textbf{\makecell{LaMP-5U \\ (Rouge-L) $\uparrow$}} & \textbf{\makecell{Required \\ VRAM $\downarrow$}}\\
    \hline
    \makecell{ChatGPT \\ \cite{Salemi2023}} & 0.658 & 0.336 & ? \\
    \hline
    \makecell{FlanT5-XXL \\ \cite{Salemi2023}} & 0.282 & \textbf{0.424} & 43 GB\\
    \hline
    \makecell{OpenChat \\ (FP16)} & 0.238 & 0.423* & 28 GB \\
    \hline
    \makecell{OpenChat \\ (INT4)} & \textbf{0.234} & 0.419* & \textbf{4.2 GB} \\
    \end{tabular}
    \end{table}

    \subsection{Benchmark comparison} Finally, we compared our findings with the RAG results from LaMP \cite{Salemi2023}. Table~\ref{table:lamp_comp} shows that OpenChat (Contriever, $k=5$) can beat FlanT5-XXL in LaMP-3U and performs very close in LaMP-5U. More importantly, its quantized counterpart has very similar results. Since we do not have the per-sample scores for the baseline models from LaMP, we perform a one-sample t-test on the Rouge-L scores. The corresponding $p$ value of 0.29 shows a non-significant difference between the results. Moreover, the results from the LaMP paper are with finetuned retrievers while our results are with non-finetuned retrievers. This indicates that a quantized-7B LLM can compete and even outperform a bigger model on personalization with RAG. 
    
    Table~\ref{table:lamp_comp} shows how much GPU VRAM is needed to deploy each model. With this comparison, the benefit of quantization becomes more pronounced: multiple high-level consumer GPUs or an A100 is necessary for running even a 7B LLM while a mid-level consumer GPU (eg. RTX 3060) would be enough to run it with quantization. According to the scores taken from LaMP, both FlanT5-XXL and OpenChat decisively beat ChatGPT, but the authors warn that the prompts used for ChatGPT may not be ideal and may contribute to a sub-optimal performance. Therefore, our results should not be used to make a comparison with ChatGPT.

\section{Discussion}

    Our results show that some LLMs (in particular OpenChat) can be successful in RAG pipelines, even after quantization, but the performance is LLM- and task-dependent. The method of quantization affects LLMs differently \cite{li2024}. Thus, the relationship between quantization and RAG performance is not straightforward and can be studied more extensively. Still, our results indicate that when a small LLM performs the task well, its AWQ-quantized counterpart performs on par. 
    
    The differing performance of some LLMs between the datasets may be partly due to prompting. LLMs are sensitive to prompts, and a prompt that works for one LLM may not work for another one \cite{Sclar2023}. The most peculiar result is the lackluster performance of LLaMA3 in LaMP-5U. LLaMA3 is a recently released model trained with an extensive pretraining corpus \cite{LLaMA3}. It has a higher chance of seeing the abstracts presented in the LaMP-5U in its pretraining data. This may explain its superior zero-shot performance. LLMs suffer from a knowledge conflict between their parametric information and the contextual information presented through retrieval \cite{xu2024knowledge}. If LLaMA3 had already memorized some of the titles of the abstracts in LaMP-5U, it might result in a knowledge conflict when similar abstract-title pairs of the same author are presented. This may explain the reduced improvement in its performance with retrieval.

    LLMs have been shown to struggle with too many retrieved documents \cite{Nelson2023}, and our findings are in accordance. Our results indicate that more than $>5$ documents do not help and can even hurt performance. From prior studies, we know that LLMs focus more on the bottom and the top of their context window \cite{Nelson2023}. We progressively put the most relevant documents starting from the bottom to the top. Therefore especially for $k=max$ settings, the less relevant documents are put on the top. This situation might hurt the LLM performance as it would focus on the most and the least related information in this case. That being said, state-of-the-art LLMs with more than 7B parameters also suffer from the same phenomenon even when not quantized \cite{Nelson2023}. Although quantization increases the risk of worsened long-context performance, we cannot conclude that it is the sole perpetrator, as this is an inherent problem for all LLMs.     

\section{Conclusion}

    We have shown that quantized smaller LLMs can use RAG to perform complex tasks such as personalization. Even though quantization might decrease the ability of LLMs to analyze long contexts, it is task- and LLM-dependent. An LLM that performs well on a task does not lose much of its long-context abilities when quantized. Thus, we conclude that quantized 7B LLMs can be the backbones of RAG with long contexts. The reduced computational load obtained from quantization would make it possible to run RAG applications with more affordable and accessible hardware. For future work, more quantization methods can be included in the experiments to see if the findings can be replicated across different methods. We can also extend the number set of k, especially between $k=5$ and $k=max\_4K$, and change the order of the documents to better understand how quantized LLMs use their context windows.

\begin{acknowledgments}
    This research is part of the project LESSEN with project number NWA.1389.20.183 of the research program NWA\_ORC 2020\/21, which is (partly) funded by the Dutch Research Council (NWO).
\end{acknowledgments}

\bibliography{main}

\end{document}